\documentclass[runningheads]{llncs}
\usepackage[T1]{fontenc}
\usepackage{graphicx}
\usepackage{hyperref}
\usepackage{tabularx}
\usepackage{amsmath}
\usepackage{amsfonts,amssymb}
\usepackage{multirow}
\usepackage{booktabs}
\usepackage{bbding}
\usepackage{ulem}
\usepackage{bm}
\usepackage{doi}
\usepackage[lined,boxed,commentsnumbered,ruled]{algorithm2e}

\begin{document}

\title{Lifelong Histopathology Whole Slide Image Retrieval via Distance Consistency Rehearsal}
\titlerunning{Lifelong Histopathology Whole Slide Image Retrieval via DCR}

\author{Xinyu Zhu \inst{1} \and Zhiguo Jiang \inst{1} \and Kun Wu \inst{1} \and Jun Shi \inst{2} \and Yushan Zheng \inst{3(}\Envelope\inst{)}}


\authorrunning{Xinyu Zhu et al.}

\institute{Image Processing Center, School of Astronautics, Beihang University, Beijing, 102206, China. \and School of Software, Hefei University of Technology, Hefei 230601, China. \and School of Engineering Medicine, Beijing Advanced Innovation Center on Biomedical Engineering, Beihang University, Beijing 100191, China.\\\email{yszheng@buaa.edu.cn} }

\maketitle

\begin{abstract}
Content-based histopathological image retrieval (CBHIR) has gained attention in recent years, offering the capability to return histopathology images that are content-wise similar to the query one from an established database. However, in clinical practice, the continuously expanding size of WSI databases limits the practical application of the current CBHIR methods. In this paper, we propose a Lifelong Whole Slide Retrieval (LWSR) framework to address the challenges of catastrophic forgetting by progressive model updating on continuously growing retrieval database. Our framework aims to achieve the balance between stability and plasticity during continuous learning. To preserve system plasticity, we utilize local memory bank with reservoir sampling method to save instances, which can comprehensively encompass the feature spaces of both old and new tasks. Furthermore, A distance consistency rehearsal (DCR) module is designed to ensure the retrieval queue's consistency for previous tasks, which is regarded as stability within a lifelong CBHIR system. We evaluated the proposed method on four public WSI datasets from TCGA projects. The experimental results have demonstrated the proposed method is effective and is superior to the state-of-the-art methods. The code is available at \url{https://github.com/OliverZXY/LWSR}.

\keywords{Histopathology image analysis \and CBHIR \and Continual learning.}
\end{abstract}

\section{Introduction} \label{introduction}
With the development of digital pathology and artificial intelligence, computer-aided cancer diagnosis methods (CAD) on histopathological image analysis have been widely studied \cite{lu2021data,chen2021pan,chen2022scaling,chen2022self}. Content-based histopathological image retrieval (CBHIR) is an emerging technology in this domain \cite{chen2022fast}, which offers the capability to return histopathology whole slide images (WSIs) that are content-wise similar to query WSIs from an established database.

However, though current CBHIR systems have shown excellent performance on stationary database \cite{zheng2020tracing,zheng2020diagnostic}. Nevertheless, they would also forget previously learned knowledge while learning new data, known as catastrophic forgetting \cite{french1999catastrophic,goodfellow2013empirical,chaudhry2018riemannian,huang2023conslide} in artificial intelligence systems. In image retrieval systems, catastrophic forgetting manifests as inconsistencies in the returned queues for old tasks before and after learning new tasks. This is rather serious for an image retrieval system in the medical scenario, because doctors may depend on the retrieved results to diagnose. Inconsistencies in these results over time can severely undermine the reliability of the system, potentially misleading doctors in diagnosis. This inevitably limits CBHIR systems’ practical application in data-incremental clinical scenario.

Recently, continual learning (CL) has been proposed \cite{lopez2017gradient,de2021continual} to mitigate catastrophic forgetting when learning continuously from a non-stationary data stream. Numerous proposed CL methods can be roughly summarized into three categories: replay methods, regularization-based methods and parameter isolation methods \cite{de2021continual}. Among them, replay methods achieve promising performance by storing a subset of data from passing data stream as exemplars for experience replay. In natural scene, lots of replay methods have been proposed to alleviate catastrophic forgetting in downstream tasks, such as classification \cite{rebuffi2017icarl,yan2021dynamically,wang2022learning} and semantic segmentation\cite{douillard2021plop,maracani2021recall}. In digital pathology, rehearsal method has been successfully applied in WSI classification task \cite{huang2023conslide}. No matter in natural scenes or in digital pathology, for the replay method, the most crucial aspect is how to sample rehearsal instances and how to replay. This holds true for digital pathology retrieval tasks as well. In the domain of CBHIR, there is a lack of research on continuous retrieval frameworks.

In this paper, we propose a novel continual whole slide retrieval framework named lifelong whole slide retrieval (LWSR). LWSR achieves good retrieval performance by balancing the trade-off between stability and plasticity during continuous learning. To preserve model's plastcity, we utilize a local memory bank to save instances of old tasks, which aims to encompass the feature space of old tasks with a reservoir random sampling algorithm \cite{vitter1985random}. Furthermore, we regard the retrieval queue's consistency for previous tasks as stability within a lifelong CBHIR system and design a distance consistency rehearsal (DCR) module, which has been verified effective to improve the performance of retrieval and the stability of returned queues after continuous learning. The proposed framework was evaluated on a public TCGA continual dataset, which contains TCGA-NSCLC, TCGA-BRCA, TCGA-RCC and TCGA-GAST. The experimental results have demonstrated the proposed LWSR is effective in the task of continual WSI retrieval and is superior to the typical classification-oriented continual learning approaches.

The contribution of the paper can be summarized in two aspects. (1) We propose a novel lifelong whole slide retrieval (LWSR) framework to tackle the challenge of continual learning in histopathology image retrieval task. To the best of our knowledge, this is the first time to solve the continual learning problem in domain of histopathology image retrieval. (2) A novel distance consistency rehearsal (DCR) module is proposed to achieve consistency of result queues for old tasks. With the local memory bank, the proposed framework can achieve balance between stability and plasticity.

\begin{figure}[t]
\includegraphics[width=\textwidth]{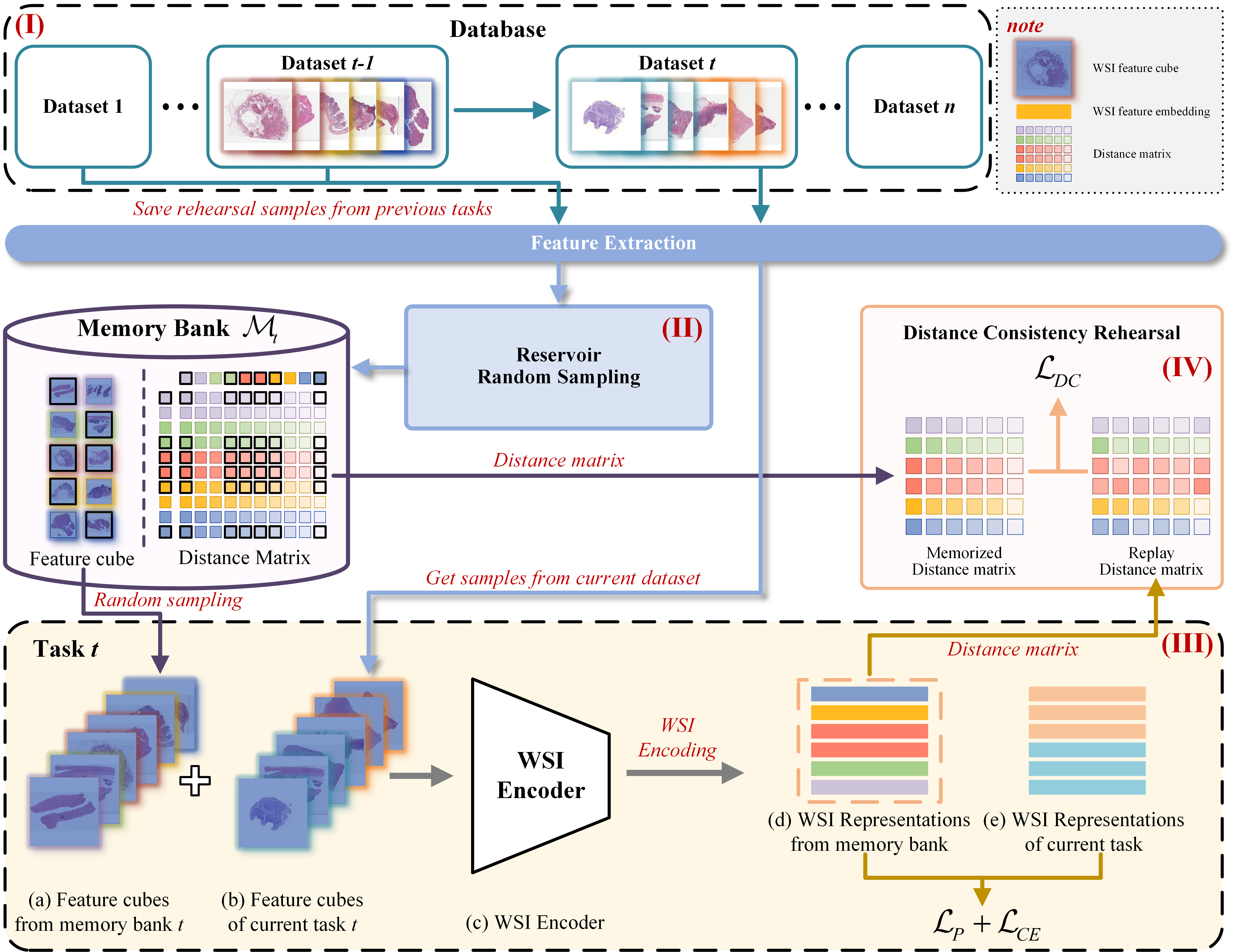}
\caption{The overview of the proposed lifelong whole slide retrieval (LWSR) framework, where (I) shows the ever-growing histopathology image database, (II) is the reservoir random sampling method for buffer sampling, (III) describe universal training process of every task, and (IV) illutrates the proposed distance consistency rehearsal (DCR) module that is detailed in section ~\ref{section_2_2} and algorithm ~\ref{algorithm_dcr}.} \label{fig1}
\end{figure}

\section{Method}
\subsection{Data Preparation and Problem Definition}
Continual learning for CBHIR can be defined as training a model on non-stationary data when new data added into the WSI database.

The proposed framework is illustrated in Fig. \ref{fig1}. For an original WSI, we first divided it into several non-overlapping patches with size of $256 \times 256$. Then, we utilize pre-trained encoder PLIP \cite{huang2023visual} to extract patch features, for its excellent performance in understanding image semantic information. Afterwards, a feature cube of a WSI can be represented as $x \in \mathbb{R}^{n_p \times d_f}$, where $d_f$ is the dimension of the feature and $n_p$ is the number of patches in the WSI. To alleviate storage pressure, we save feature cubes in the memory bank instead of the original WSIs.

As shown in Fig. \ref{fig1}(I), we define the database as a sequence of datasets $\mathcal{D} = \{ \mathcal{D}_{1}, \cdots, \mathcal{D}_{T}\}$, where $\mathcal{D}_{t} = \{ x_{i}, y_{i} \}^{N_{t}}_{i=1}$ is the dataset of task $t$, and $T$ is the total number of tasks. Dataset $\mathcal{D}_{t}$ contains $N_{t}$ labeled WSIs, and $y_{i}$ is the class label of WSI $x_{i}$. In the class-incremental CHBIR scenario, the model needs to return relative WSIs sequence of given query WSIs from both old tasks and new tasks. Universal training process for every task is illustrated in Fig. \ref{fig1}(III). When data of new task coming, described as Fig. \ref{fig1}(a-b), a mini-batch of training data and replay samples from memory bank are input to the WSI encoder to get WSI representations. Then, representations of current task and previous tasks are combined to calculate retrieval relative losses which facilitate retrieval precision of the model. Meanwhile, we execute distance consistency rehearsal on representations of old tasks to maintain the stability of the returned queues, as shown in Fig. \ref{fig1}(IV).

\subsection{Distance Consistency Rehearsal}\label{section_2_2}
As illustrated in section \ref{introduction}, we design distance consistency rehearsal (DCR) module to maintain stability of result queues for old tasks, which is detailed in Fig. \ref{fig1}(IV) and illustrated in algorithm \ref{algorithm_dcr}.

In practical scenarios, the stability of the CBHIR system is demonstrated by the consistency of the result queues for old tasks. In terms of feature space, this stability is reflected by maintaining constant distances between instances of old tasks even after learning new tasks. Based on this prior, we propose the distance consistency rehearsal (DCR), which is achieved by minimizing the differential value of distance between representations of current replay samples and distance matrix of corresponding replay samples saved in memory bank. The rehearsal process is elaborated as follows.

First, we construct the target distance matrix of instances saved in memory bank. At the ending of task $t-1$, we execute reservoir sampling algorithm \cite{vitter1985random} on dataset $\mathcal{D}_{t-1}$ and memory bank $\mathcal{M}_{t-1}$ to get sampled feature cubes $\mathbf{X}_{t-1} \in \mathbb{R}^{n_{t-1} \times n_{p} \times d_{f}}$ and $\mathbf{X}_{t-1}^{m} \in \mathbb{R}^{n_{t-1}^{m} \times n_{p} \times d_{f}}$ for the memory bank $\mathcal{M}_{t}$ of task $t$, where $n_{t-1}$ and $n_{t-1}^{m}$ means sampled number of $\mathcal{D}_{t-1}$ and $\mathcal{M}_{t-1}$ respectively. Then, sampled feature cubes is fed into task $t-1$ encoder $f_{t-1}$ to get feature representations $\mathbf{F}_{t-1} \in \mathbb{R}^{n_{t-1} \times d_{F}}$ and $\mathbf{F}_{t-1}^{m} \in \mathbb{R}^{n_{t-1}^{m} \times d_{F}}$, where $d_{F}$ is the dimension of the whole slide representation. With combination of $\mathbf{F}_{t-1}$ and $\mathbf{F}_{t-1}^{m}$, denoted as $\mathbf{C}_{t-1} \in \mathbb{R}^{n_{t-1}^{d} \times d_{F}}$, we calculate the Euclidean distance between every pair of elements in set $\mathbf{C}_{t-1}$ to obtain the target distance matrix $\mathbf{D}_{t-1}$, which is formulated as
\begin{equation} \label{eq1}
\mathbf{D}_{t-1}\left[i,j\right] = d\left( \mathbf{c}_{t-1}^{i},\mathbf{c}_{t-1}^{j} \right) = \|\mathbf{c}_{t-1}^{i}-\mathbf{c}_{t-1}^{j}\|_2
\end{equation}
where $\mathbf{D}_{t-1} \in \mathbb{R}^{n_{t-1}^{d} \times n_{t-1}}$, $n_{t-1}^{d}=n_{t-1}+n_{t-1}^{m}$, denotes the total number of instances sampled at the ending of task $t-1$, and $\mathbf{c}_{t-1}^{i}, \mathbf{c}_{t-1}^{j}$ represent the $i$-th and $j$-th sampled instance.

Then, distance consistency rehearsal is implemented for current task. In every mini-batch of current dataset $\mathcal{D}_{t}$, we get representations of current dataset and instances sampled from memory bank, denoted as $\mathbf{F}_{t}$ and $\mathbf{F}_{t}^{m}$. For $\mathbf{F}_{t}^{m}$, distance matrix $\mathbf{d}_{t}$ is obtained by the same way in Eq. (\ref{eq1}) and illustrated in algorithm \ref{algorithm_dcr}. To maintain the distance consistency of rehearsal samples, we minimize the Mean Squared Error (MSE) loss between distance matrix of current task and replay samples by the equations
\begin{equation} \label{eq2}
\mathcal{L}_{DC} \left( \mathbf{d}_{t}, \mathbf{d}_{t-1} \right) = \|\mathbf{d}_{t}-\mathbf{d}_{t-1}\|_F^2
\end{equation}
where $n_{t}^{m}$ means total number of feature cubes sampled from memory bank and $\mathbf{d}_{t-1}$ is sub-matrix sampled from $\mathbf{D}_{t-1}$ according to the indexes of $\mathbf{X}_{t}^{m}$.

\begin{algorithm}\scriptsize
\label{algorithm_dcr}
\caption{Distance Consistency Rehearsal algorithm.}
\KwIn{current encoder $f_{t}$, previous encoder$f_{t-1}$, current dataset $\mathcal{D}_{t}$, previous dataset $\mathcal{D}_{t-1}$, current memory bank $\mathcal{M}_{t}$ and previous memory bank $\mathcal{M}_{t-1}$}
\tcp{Target distance matrix construction}
$\mathcal{M}_{t}\leftarrow{\mathbf{X}_{t-1}, \mathbf{X}_{t-1}^{m}}\leftarrow$ RESERVOIR$\left( \mathcal{D}_{t-1}, \mathcal{M}_{t-1} \right)$

$\mathbf{F}_{t-1}, \mathbf{F}_{t-1}^{m}\leftarrow f_{t-1} \left( \mathbf{X}_{t-1} \right), f_{t-1} \left( \mathbf{X}_{t-1}^{m} \right)$

$\mathbf{C}_{t-1}\leftarrow$ Concat$\left( \mathbf{F}_{t-1}, \mathbf{F}_{t-1}^{m} \right)$

\For{$\mathbf{c}_{t-1}^{i}, \mathbf{c}_{t-1}^{j}$ in $\mathbf{C}_{t-1}$}{
$\mathbf{D}_{t-1}\left[i,j\right] \leftarrow d\left( \mathbf{c}_{t-1}^{i},\mathbf{c}_{t-1}^{j} \right) = \|\mathbf{c}_{t-1}^{i}-\mathbf{c}_{t-1}^{j}\|_2$
}

\tcp{Distance consistency rehearsal for current task}

\For{each mini-batch $\mathbf{X}_{t}$ in $\mathcal{D}_{t}$}{
\tcp{$\mathbf{d}_{t-1}$ is the the sub-matrix sampled from $\mathbf{D}_{t-1}$ according to the indexes of $\mathbf{X}_{t}^{m}$}
$\mathbf{X}_{t}^{m}, \mathbf{d}_{t-1} \leftarrow$ Random Sampling$\left( \mathcal{M}_{t} \right)$

$\mathbf{F}_{t}, \mathbf{F}_{t}^{m}\leftarrow f_{t}\left(\mathbf{X}_{t}\right), f_{t}\left(\mathbf{X}_{t}^{m}\right)$

\For{$\mathbf{f}_{t}^{i}, \mathbf{f}_{t}^{j}$ in $\mathbf{F}_{t}^{m}$}{
$\mathbf{d}_{t}\left[i,j\right] \leftarrow d\left( \mathbf{f}_{t}^{i},\mathbf{f}_{t}^{j} \right) = \|\mathbf{f}_{t}^{i}-\mathbf{f}_{t}^{j}\|_2$
}

$\mathcal{L}_{DC} \left( \mathbf{d}_{t}, \mathbf{d}_{t-1} \right) = \|\mathbf{d}_{t}-\mathbf{d}_{t-1}\|_F^2$

}

\end{algorithm}

\subsection{Objective for LWSR}
Above all, we design the distance consistency loss $\mathcal{L}_{DC}$ to maintain the queue stability of a continual CBHIR system. Meanwhile, pair-wise loss and cross-entropy loss have been proven effective in image retrieval task \cite{zheng2020diagnostic,zheng2020tracing}. As a result, the model is optimized in an end-to-end fusion by
\begin{equation} \label{eq3}
\mathcal{L}_{all} = \mathcal{L}_{P} \left( \mathbf{F}_{t}, \mathbf{F}_{t}^{m}, y_t \right) + \mathcal{L}_{CE} \left( y_t, y_t^{pred} \right) + \alpha \mathcal{L}_{DC} \left( \mathbf{d}_{t}, \mathbf{d}_{t-1} \right),
\end{equation}
where $\mathcal{L}_{P}$ is the pair-wise loss, $\mathcal{L}_{CE}$ is the cross-entropy loss, $\mathcal{L}_{DC}$ is the distance consistency loss, $y_t$ and $y_t^{pred}$ are the ground-truth label and output logits of input instances, and $\alpha$ is a balancing factor. Through \ref{eq3}, we attempt to not only improve retrieval precision of LWSR but also maintain queue consistency as much as possible.

\section{Experiment and Result}
To comprehensively evaluate our proposed method, we construct a sequence WSI retrieval dataset, details illustrated in Table. \ref{tab_dataset}. The sequence dataset contains four datasets from The Cancer Genome Atlas (TCGA), which is listed as non-small cell lung carcinoma (NSCLC),  renal cell carcinoma (RCC), invasive breast carcinoma (BRCA) and gastrointestinal adenocarcinoma (GAST).
We define the continual histopathology image database as a dynamic growing database. Each time one dataset containing two classes of cancer is added into the database, which could be seen as a class-incremental retrieval scene. The arrival sequence of continual database is NSCLC, RCC, BRCA and GAST.

We used pre-trained encoder PLIP \cite{huang2023visual} to extract patch features on the magnification under 20× lenses. TransMIL \cite{shao2021transmil} was utilized as the WSI encoder (shown in Fig. \ref{fig1}(c)) for its effectiveness has been verified on various WSI analysis tasks \cite{shao2021transmil,wang2022transformer}. All the methods were implemented in Python 3.8 with PyTorch 1.12 and CUDA 12.2 and run the experiments on a computer with 2 GPUs of Nvidia Geforce RTX 4090.

For evaluation of retrieval precision, we report six metrics, containing slide-level mAP, R@3, and P@5 to evaluate the performance of 8-class tumor type retrieval, and site-mAP, site-R@3, site-P@5 for the 4-anatomic-site retrieval. Furthermore, to evaluate the stability of our method, we adopt two statistical correlation metrics Spearman's rank correlation coefficient (SRC) \cite{spearman1987proof} and Kendall rank correlation coefficient (KRC) \cite{kendall1938new} to assess the consistency of returned queues for old tasks.

\begin{table}[t]\scriptsize
		\centering
		\caption{Illustration of sequence TCGA dataset.}
		\label{tab_dataset}
		\setlength{\tabcolsep}{4pt}
		\begin{tabular}{ccccc}
			\toprule
                \textbf{Dataset} & \textbf{Anatomic site} & \textbf{Tumor type} & \textbf{Cases} & \textbf{Total Cases} \\
                \midrule
                \multirow{2}{*}{NSCLC} & 
                \multirow{2}{*}{Pulmonary} & 
                Lung adenocarcinoma (LUAD) & 
                496 & 
                \multirow{2}{*}{995} \\
                & & Lung squamous cell carcinoma (LUSC) & 499 \\
                \midrule
                \multirow{2}{*}{RCC} & 
                \multirow{2}{*}{Urinary} & 
                Clear cell renal cell carcinoma (KIRC) & 
                492 & 
                \multirow{2}{*}{779} \\
                & & Papillary renal cell carcinoma (KIRP) & 287 \\
                \midrule
                \multirow{2}{*}{BRCA} & 
                \multirow{2}{*}{Breast} & 
                Invasive ductal (IDC) & 
                794 & 
                \multirow{2}{*}{998} \\
                & & Invasive lobular carcinoma (ILC) & 204 \\
                \midrule
                \multirow{2}{*}{GAST} & 
                \multirow{2}{*}{Gastrointestinal} & 
                Colon Adenocarcinoma (COAD) & 
                369 & 
                \multirow{2}{*}{731} \\
                & & Stomach Adenocarcinoma (STAD) & 362 \\
                \midrule \midrule
                \textbf{Total Count} & & & & \textbf{3053} \\
			\bottomrule
		\end{tabular}
	\end{table}

\subsection{Model verification}
\begin{figure}[!t]
\includegraphics[width=\textwidth]{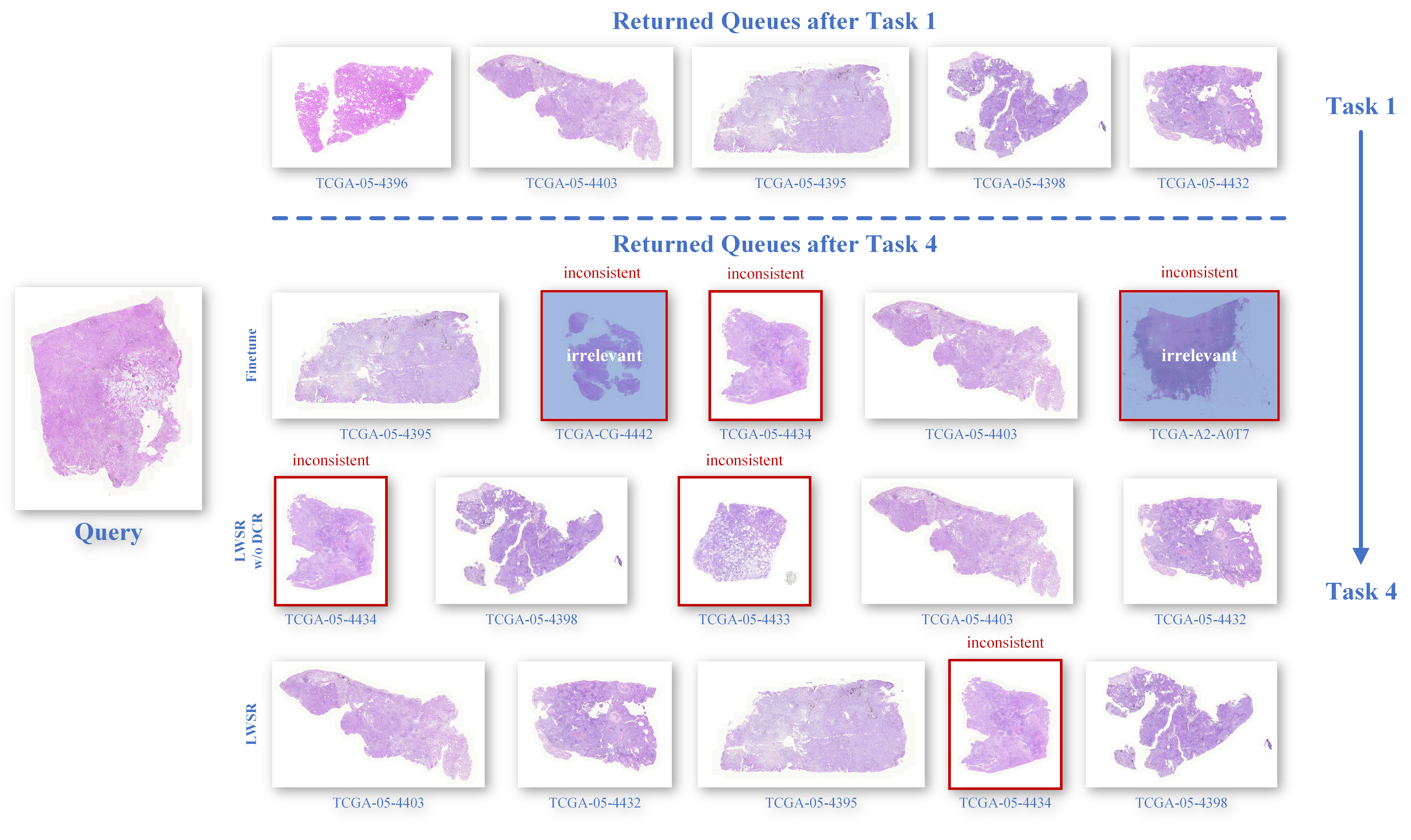}
\caption{Visualization of the capacity of the proposed method in keeping returned queues consistent after continual learning, inconsistent returned WSIs are framed in red and irrelevant results are filled with blue.} \label{fig2}
\end{figure}

We first trained a model with all datasets under fully supervision as the upper-bound, which is identified as \textit{JointTrain} in Table. \ref{tab1}. Subsequently, individual training were sequentially conducted for the four datasets as the lower boundary, which is identified as \textit{Finetune}. According to Table. \ref{tab1}, model subjected to \textit{Finetune} exhibits the poorest performance, indicating that catastrophic forgetting occurs as the database size increases. In contrast, our method significantly mitigates catastrophic forgetting, performing nearly as well as the \textit{JointTrain} approach. Although our method does not achieve as high mAP as \textit{JointTrain}, it delivers comparable values in R@3 and P@5 metrics with \textit{JointTrain}, which means the pathologists can efficiently receive diagnostically relevant cases by assessing a few top-returned results. DCR is defined to maintain the distance relationship between the cases from old retrieval tasks, which provides more finer guidance to the model in learning the similarities between WSIs compared to the common pair-wise loss. As highlighted in Table. \ref{tab1} and illustrated in Fig. \ref{fig2}, DCR markedly improves the precision and consistency of fine-grained retrieval, as demonstrated by enhancements in both the precision and consistency metrics. Moreover, we combined Breakup-Reorganize (BuRo) rehearsal, the main component of ConSlide \cite{huang2023conslide}, with our framework, which is presented as LWSR w/ BuRo in Table. \ref{tab1}. BuRo achieves data augmentation by randomly selecting and combining patches from same category WSIs. However, for retrieval tasks, representations are expected to describe the complete semantic information of WSIs. The virtual cases created by BuRo would mislead the retrieval model to describe the real WSIs. It can be obviously observed that BuRo rehearsal method performs poorly in WSI retrieval tasks, with mAP of 0.563, showing a significant performance gap between LWSR w/o DCR and LWSR.

\begin{table}[t]\scriptsize
		\centering
		\caption{Comparison with baselines and ablation study.}
		\label{tab1}
		\setlength{\tabcolsep}{4pt}
            \resizebox{\textwidth}{!}{
		\begin{tabular}{l|cccccc|cc|}
			\toprule
			\multicolumn{1}{l|}{\multirow{2}{*}{\textbf{Methods}}} & \multicolumn{6}{c|}{\textbf{Retrieval Precision Metric}} & \multicolumn{2}{c}{\textbf{Consistency Metric}}  \\ 
			\multicolumn{1}{c|}{} & \multicolumn{1}{c}{mAP} & \multicolumn{1}{c}{R@3} & \multicolumn{1}{c|}{P@5} & \multicolumn{1}{c}{site mAP} & \multicolumn{1}{c}{site R@3} & \multicolumn{1}{c|}{site P@5} & \multicolumn{1}{c}{SRC} & \multicolumn{1}{c}{KRC} \\ 
		\midrule
                JointTrain &
                \multicolumn{1}{c}{\bfseries\underline{0.908}} & 
                \multicolumn{1}{c}{0.914} &
			    \multicolumn{1}{c|}{\bfseries\underline{0.903}} &
		      \multicolumn{1}{c}{\bfseries\underline{0.833}} &
                \multicolumn{1}{c}{0.987} &
			    \multicolumn{1}{c|}{0.983} &
                \multicolumn{1}{c}{-} &
                \multicolumn{1}{c}{-}\\
                Finetune &
                \multicolumn{1}{c}{0.426} & 
                \multicolumn{1}{c}{0.877} &
			    \multicolumn{1}{c|}{0.635} &
		      \multicolumn{1}{c}{0.557} &
                \multicolumn{1}{c}{0.965} &
			    \multicolumn{1}{c|}{0.824} &
                \multicolumn{1}{c}{0.408} &
                \multicolumn{1}{c}{0.295}\\
                \textbf{LWSR w/ BuRo} &
                \multicolumn{1}{c}{0.563} & 
                \multicolumn{1}{c}{0.940} &
			    \multicolumn{1}{c|}{0.809} &
		      \multicolumn{1}{c}{0.628} &
                \multicolumn{1}{c}{0.988} &
			    \multicolumn{1}{c|}{0.948} &
                \multicolumn{1}{c}{0.687} &
                \multicolumn{1}{c}{0.535}\\
                \textbf{LWSR w/o DCR} &
                \multicolumn{1}{c}{0.789} & 
                \multicolumn{1}{c}{0.938} &
			    \multicolumn{1}{c|}{0.861} &
		      \multicolumn{1}{c}{0.757} &
                \multicolumn{1}{c}{0.988} &
			    \multicolumn{1}{c|}{0.974} &
                \multicolumn{1}{c}{0.852} &
                \multicolumn{1}{c}{0.707}\\
                \textbf{LWSR} &
                \multicolumn{1}{c}{0.796} & 
                \multicolumn{1}{c}{\bfseries\underline{0.960}} &
			    \multicolumn{1}{c|}{0.871} &
		      \multicolumn{1}{c}{0.804} &
                \multicolumn{1}{c}{\bfseries\underline{0.997}} &
			    \multicolumn{1}{c|}{\bfseries\underline{0.990}} &
                \multicolumn{1}{c}{\bfseries\underline{0.878}} &
                \multicolumn{1}{c}{\bfseries\underline{0.736}}\\
			\bottomrule
		\end{tabular}}
	\end{table}

\subsection{Comparison with other continuous learning frameworks}
We compare our method with several classic continual learning approaches, involving regularization-based methods LwF \cite{li2017learning} and EWC \cite{kirkpatrick2017overcoming} and replay-based methods ER-ACE \cite{caccia2021new}, A-GEM \cite{chaudhry2018riemannian} and DER++ \cite{buzzega2020dark}.

\begin{table}[t]
\scriptsize
		\centering
		\caption{Comparison with other continual learning methods.}
		\label{tab2}
		\setlength{\tabcolsep}{4pt}
            \resizebox{\textwidth}{!}{
		\begin{tabular}{l|cccccc|cc|}
			\toprule
			\multicolumn{1}{l|}{\multirow{2}{*}{\textbf{Methods}}} & \multicolumn{6}{c|}{\textbf{Retrieval Precision Metric}} & \multicolumn{2}{c}{\textbf{Consistency Metric}}  \\ 
			\multicolumn{1}{c|}{} & \multicolumn{1}{c}{mAP} & \multicolumn{1}{c}{R@3} & \multicolumn{1}{c|}{P@5} & \multicolumn{1}{c}{site mAP} & \multicolumn{1}{c}{site R@3} & \multicolumn{1}{c|}{site P@5} & \multicolumn{1}{c}{SRC} & \multicolumn{1}{c}{KRC} \\ 
            \midrule
            \multicolumn{9}{c}{\textbf{Regularization-based}} \\
            \midrule
                LwF\cite{li2017learning} &
                \multicolumn{1}{c}{0.432} & 
                \multicolumn{1}{c}{0.895} &
			    \multicolumn{1}{c|}{0.647} &
		      \multicolumn{1}{c}{0.560} &
                \multicolumn{1}{c}{0.965} &
			    \multicolumn{1}{c|}{0.829} &
                \multicolumn{1}{c}{0.473} &
                \multicolumn{1}{c}{0.339}\\
                EWC\cite{kirkpatrick2017overcoming} &
                \multicolumn{1}{c}{0.437} & 
                \multicolumn{1}{c}{0.881} &
			    \multicolumn{1}{c|}{0.647} &
		      \multicolumn{1}{c}{0.573} &
                \multicolumn{1}{c}{0.970} &
			    \multicolumn{1}{c|}{0.836} &
                \multicolumn{1}{c}{0.440} &
                \multicolumn{1}{c}{0.319}\\
            \midrule
            \multicolumn{9}{c}{\textbf{Replay based (Buffer size $\approx$ 5 WSIs)}} \\
            \midrule
                ER-ACE\cite{caccia2021new} &
                \multicolumn{1}{c}{0.672} & 
                \multicolumn{1}{c}{0.931} &
			    \multicolumn{1}{c|}{0.815} &
		      \multicolumn{1}{c}{0.685} &
                \multicolumn{1}{c}{0.985} &
			    \multicolumn{1}{c|}{0.954} &
                \multicolumn{1}{c}{0.818} &
                \multicolumn{1}{c}{0.687}\\
                A-GEM\cite{chaudhry2018efficient} &
                \multicolumn{1}{c}{0.439} & 
                \multicolumn{1}{c}{0.897} &
			    \multicolumn{1}{c|}{0.687} &
		      \multicolumn{1}{c}{0.597} &
                \multicolumn{1}{c}{0.972} &
			    \multicolumn{1}{c|}{0.850} &
                \multicolumn{1}{c}{0.659} &
                \multicolumn{1}{c}{0.507}\\
                DER++\cite{buzzega2020dark} &
                \multicolumn{1}{c}{0.655} & 
                \multicolumn{1}{c}{0.910} &
			    \multicolumn{1}{c|}{0.786} &
		      \multicolumn{1}{c}{0.728} &
                \multicolumn{1}{c}{0.985} &
			    \multicolumn{1}{c|}{0.948} &
                \multicolumn{1}{c}{0.835} &
                \multicolumn{1}{c}{0.696}\\
                \textbf{LWSR} &
                \multicolumn{1}{c}{\bfseries\underline{0.773}} & 
                \multicolumn{1}{c}{\bfseries\underline{0.945}} &
			    \multicolumn{1}{c|}{\bfseries\underline{0.859}} &
		      \multicolumn{1}{c}{\bfseries\underline{0.788}} &
                \multicolumn{1}{c}{\bfseries\underline{0.998}} &
			    \multicolumn{1}{c|}{\bfseries\underline{0.979}} &
                \multicolumn{1}{c}{\bfseries\underline{0.863}} &
                \multicolumn{1}{c}{\bfseries\underline{0.723}}\\
            \midrule
            \multicolumn{9}{c}{\textbf{Replay based (Buffer size $\approx$ 10 WSIs)}} \\
            \midrule
                ER-ACE\cite{caccia2021new} &
                \multicolumn{1}{c}{0.733} & 
                \multicolumn{1}{c}{0.930} &
			    \multicolumn{1}{c|}{0.836} &
		      \multicolumn{1}{c}{0.761} &
                \multicolumn{1}{c}{0.981} &
			    \multicolumn{1}{c|}{0.964} &
                \multicolumn{1}{c}{0.854} &
                \multicolumn{1}{c}{0.716}\\
                A-GEM\cite{chaudhry2018efficient} &
                \multicolumn{1}{c}{0.496} & 
                \multicolumn{1}{c}{0.905} &
			    \multicolumn{1}{c|}{0.711} &
		      \multicolumn{1}{c}{0.652} &
                \multicolumn{1}{c}{0.971} &
			    \multicolumn{1}{c|}{0.882} &
                \multicolumn{1}{c}{0.635} &
                \multicolumn{1}{c}{0.510}\\
                DER++\cite{buzzega2020dark} &
                \multicolumn{1}{c}{0.680} & 
                \multicolumn{1}{c}{0.927} &
			    \multicolumn{1}{c|}{0.799} &
		      \multicolumn{1}{c}{0.749} &
                \multicolumn{1}{c}{0.987} &
			    \multicolumn{1}{c|}{0.949} &
                \multicolumn{1}{c}{0.839} &
                \multicolumn{1}{c}{0.695}\\
                \textbf{LWSR} &
                \multicolumn{1}{c}{\bfseries\underline{0.796}} & 
                \multicolumn{1}{c}{\bfseries\underline{0.960}} &
			    \multicolumn{1}{c|}{\bfseries\underline{0.871}} &
		      \multicolumn{1}{c}{\bfseries\underline{0.804}} &
                \multicolumn{1}{c}{\bfseries\underline{0.997}} &
			    \multicolumn{1}{c|}{\bfseries\underline{0.990}} &
                \multicolumn{1}{c}{\bfseries\underline{0.878}} &
                \multicolumn{1}{c}{\bfseries\underline{0.736}}\\
            \midrule
            \multicolumn{9}{c}{\textbf{Replay based (Buffer size $\approx$ 15 WSIs)}} \\
            \midrule
                ER-ACE\cite{caccia2021new} &
                \multicolumn{1}{c}{0.746} & 
                \multicolumn{1}{c}{0.925} &
			    \multicolumn{1}{c|}{0.835} &
		      \multicolumn{1}{c}{0.732} &
                \multicolumn{1}{c}{0.985} &
			    \multicolumn{1}{c|}{0.955} &
                \multicolumn{1}{c}{0.849} &
                \multicolumn{1}{c}{0.705}\\
                A-GEM\cite{chaudhry2018efficient} &
                \multicolumn{1}{c}{0.489} & 
                \multicolumn{1}{c}{0.887} &
			    \multicolumn{1}{c|}{0.696} &
		      \multicolumn{1}{c}{0.668} &
                \multicolumn{1}{c}{0.972} &
			    \multicolumn{1}{c|}{0.873} &
                \multicolumn{1}{c}{0.707} &
                \multicolumn{1}{c}{0.563}\\
                DER++\cite{buzzega2020dark} &
                \multicolumn{1}{c}{0.693} & 
                \multicolumn{1}{c}{0.924} &
			    \multicolumn{1}{c|}{0.799} &
		      \multicolumn{1}{c}{0.739} &
                \multicolumn{1}{c}{0.985} &
			    \multicolumn{1}{c|}{0.943} &
                \multicolumn{1}{c}{0.852} &
                \multicolumn{1}{c}{0.714}\\
                \textbf{LWSR} &
                \multicolumn{1}{c}{\bfseries\underline{0.808}} & 
                \multicolumn{1}{c}{\bfseries\underline{0.931}} &
			    \multicolumn{1}{c|}{\bfseries\underline{0.855}} &
		      \multicolumn{1}{c}{\bfseries\underline{0.826}} &
                \multicolumn{1}{c}{\bfseries\underline{0.992}} &
			    \multicolumn{1}{c|}{\bfseries\underline{0.983}} &
                \multicolumn{1}{c}{\bfseries\underline{0.866}} &
                \multicolumn{1}{c}{\bfseries\underline{0.715}}\\
			\bottomrule
		\end{tabular}}
	\end{table}

According to Table. \ref{tab2}, regularization based methods fail to effectively alleviate catastrophic forgetting and they only achieve a slight improvement over \textit{Finetune}. Compared to other replay-based continual learning methods, our method consistently achieve the best performance under various circumstances. The main reason is that these methods are proposed to solve catastrophic forgetting in the domain of classification initially and do not accounted for taking the interaction between the current task and previous tasks into consideration. In comparison with the second-best methods under buffer in size of about 10 WSIs, LWSR achieves an increase of 6.3\%/3.0\%/3.5\% in mAP, R@3 and P@5, and an increase of 4.3\%/1.6\%/2.6\% in site mAP, R@3 and P@5. It has demonstrated the effectiveness of the proposed method for the task of lifelong CBHIR. Furthermore, with the growth of buffer size, our method can achieve further improvements. It is also apparent that the enhancement in performance metrics upon expanding the buffer size from 10 WSIs to 15 WSIs is not as pronounced as the improvement observed from enlarging it from 5 to 10. A 10-WSI buffer balances diversity between current and previous tasks, aiding the model in achieving high retrieval precision after sequential learning. However, a 15-WSI buffer may overly emphasize past tasks at the expense of current ones, leading to decreased focus on the present task.

\section{Conclusion}
In this paper, we have proposed a novel Lifelong Whole Slide Retrieval (LWSR) framework to address the challenges of catastrophic forgetting by progressive model updating on continuously growing retrieval database. The experiments on a large sequence retrieval dataset have proven the LWSR is effective in the scenario of class-incremental retrieval tasks, and achieves the state-of-the-art overall performance.

\begin{credits}
\subsubsection{\ackname} This work was partly supported by Beijing Natural Science Foundation (Grant No. 7242270), partly supported by the National Natural Science Foundation of China (Grant No. 62171007, 61901018, and 61906058), and partly supported by the Fundamental Research Funds for the Central Universities of China (grant No. YWF-23-Q-1075 and JZ2022HGTB0285).

\subsubsection{\discintname}
The authors have no competing interests to declare that are relevant to the content of this article.
\end{credits}

\newpage

\title{Supplementary Material of Lifelong Histopathology Whole Slide Image Retrieval via Distance Consistency Rehearsal}
\titlerunning{Lifelong Histopathology Whole Slide Image Retrieval via DCR}

\author{Xinyu Zhu \inst{1} \and Zhiguo Jiang \inst{1} \and Kun Wu \inst{1} \and Jun Shi \inst{2} \and Yushan Zheng \inst{3(}\Envelope\inst{)}}


\authorrunning{Xinyu Zhu et al.}

\institute{Image Processing Center, School of Astronautics, Beihang University, Beijing, 102206, China. \and School of Software, Hefei University of Technology, Hefei 230601, China. \and School of Engineering Medicine, Beijing Advanced Innovation Center on Biomedical Engineering, Beihang University, Beijing 100191, China.\\\email{yszheng@buaa.edu.cn} }

\maketitle

\section{Evaluation metrics}

\begin{table}[h!]
\caption{Metrics used in evaluation of returned queue consistency}
\centering
\begin{tabularx}{\textwidth}{| >{\raggedright\arraybackslash}X | >{\raggedright\arraybackslash}X | >{\raggedright\arraybackslash}X |}
\hline
\textbf{Metric Name} & \textbf{Definition} & \textbf{Description} \\ 
\hline
Spearman's Rank Correlation Coefficient & 
\[\rho_{i,j} = 1 - \frac{6 \sum d_k^2}{n (n^2 - 1)}\] & 
\(d_k\) is the difference between the ranks of retrieval sequence of $i$-th task after $j$ tasks' training, and \(n\) is the number of instance of each sequence. \\ 
\hline
\textbf{Usage in Application} & 
\multicolumn{2}{>{\raggedright\arraybackslash}X|}{
\[
SRC = \frac{1}{n-1} \sum_{i=1}^{n-1} \left( \frac{1}{n-i} \sum_{j=i+1}^{n} \rho_{i,j} \right)
\]
} \\ 
\hline
Kendall's Rank Correlation Coefficient & 
\[\tau_{i,j} = \frac{C - D}{\frac{n(n-1)}{2}}\] & 
\(C\) is the number of concordant pairs of retrieval sequence of $i$-th task after $j$ tasks' training, \(D\) is the number of discordant pairs, and \(n\) is the number of instance of each sequence. \\ 
\hline
\textbf{Usage in Application} & 
\multicolumn{2}{>{\raggedright\arraybackslash}X|}{
\[
KRC = \frac{1}{n-1} \sum_{i=1}^{n-1} \left( \frac{1}{n-i} \sum_{j=i+1}^{n} \tau_{i,j} \right)
\]
} \\ 
\hline
\end{tabularx}
\label{table:metrics}
\end{table}

\newpage

\section{Implementation details}

\begin{table}[h]
\caption{Hyperparameters and corresponding values.}
\centering
\begin{tabular}{ll}
\toprule
\textbf{Hyperparameter Name} & \textbf{Hyperparameter Value} \\
\midrule
Patch Sampling Number per WSI     & 2048                      \\
Pair-wise Loss Weight             & 1.0                       \\
Cross-entropy Loss Weight         & 1.0                       \\
Distance Consistency Loss Weight  & 0.01                      \\
Learning Rate                     & 1e-5                      \\
Buffer Size                       & 100, 200, 300             \\
Batch Size                        & 10                        \\
Minibatch Size                    & 30                        \\
Number of Epochs                  & 70                        \\
Optimizer                         & Adam                      \\
Scheduler                         & StepLR                    \\
\bottomrule
\end{tabular}
\label{tab:hyperparameters}
\end{table}

\end{document}